\documentclass[10pt]{article} % For LaTeX2e
\usepackage[preprint]{tmlr}
\usepackage{enumitem}
\usepackage{amsmath,amsfonts,bm}

\def\eqref#1{equation~\ref{#1}}
\def\1{\bm{1}}

\DeclareMathAlphabet{\mathsfit}{\encodingdefault}{\sfdefault}{m}{sl}
\SetMathAlphabet{\mathsfit}{bold}{\encodingdefault}{\sfdefault}{bx}{n}

\usepackage{booktabs}
\usepackage{hyperref}
\usepackage{url}

\usepackage{amsthm}   % For Theorem environments
\usepackage{amsmath}  % For additional math commands
\usepackage{amssymb}  % For additional math symbols

\newcommand{\EmpSys}{\mathcal{E}}          % empirical system
\newcommand{\NumSys}{\mathcal{N}}          % numerical system
\newcommand{\UniqGrp}{\mathcal{G}_{\!\NumSys}} % uniqueness group

\theoremstyle{definition}

\setcitestyle{numbers}
\usepackage{mathptmx}       % Times New Roman-like font
\usepackage{graphicx}       % For graphics inclusion
\usepackage{braket}         % For quantum-related operations
\usepackage{algorithm}      % For algorithms
\usepackage{algorithmicx}   % For algorithmic pseudocode
\usepackage{algpseudocode}  % For pseudocode
\usepackage{subfig}         % For sub-figures
\usepackage{tabularx}       % For advanced tables
\usepackage{array}          % For table formatting
\usepackage{bbm}            % For mathematical symbols
\usepackage{thmtools}       % For theorem tools
\usepackage{thm-restate}    % For restating theorems
\usepackage[nottoc]{tocbibind} % Add references and TOC to ToC
\usepackage{color, soul}          % For coloring text
\usepackage{multirow}       % For multirow tables
\usepackage{enumitem}
\usepackage{color,soul}
\newcommand{\Real}{\mathbb{R}}

\title{Towards Measurement Theory for Artificial Intelligence \\(Extended Abstract)}

\author{\name Elija Perrier \email elija.perrier@gmail.com \\
      \addr Centre for Quantum Software \& Information, UTS, Sydney
}

\def\month{MM}  % Insert correct month for camera-ready version
\def\year{YYYY} % Insert correct year for camera-ready version
\def\openreview{\url{https://openreview.net/forum?id=XXXX}} % Insert correct link to OpenReview for camera-ready version

\begin{document}

\maketitle

\begin{abstract}
We motivate and outline a programme for a formal theory of measurement of artificial intelligence. We argue that formalising measurement for AI will allow researchers, practitioners, and regulators to: (i) make comparisons between systems and the evaluation methods applied to them; (ii) connect frontier AI evaluations with established quantitative risk‑analysis techniques drawn from engineering and safety science; and (iii) foreground how what counts as AI capability is contingent upon the measurement operations and scales we elect to use.  We sketch a layered measurement stack, distinguish direct from indirect observables, and signpost how these ingredients provide a pathway toward a unified, calibratable taxonomy of AI phenomena.

\vspace{0.5em}
\end{abstract}

% \tableofcontents

\section{Introduction}
Evaluation of artificial intelligence is a cornerstone of developing approaches to technical and theoretical alignment research. Driven by concern over emerging AI capabilities and their potential risks, evaluation research has led to a renaissance of proposals, methods and techniques for testing and benchmarking. Yet despite the gold-rush in evaluation techniques, there is little, if any, in the way of attempts to formally establish the theoretical underpinnings of the measurement of AI specifically. Most evaluation and benchmark research is narrowly focused, based upon testing specific models against specific benchmarks, or seeking to yield outcomes in response to limited perturbations or probing of particular models. The discipline as a whole lacks the scope, rigour and depth of measurement practice characteristic of other sciences. AI evaluation rarely engages with modern measurement theory, has little in the way of metrology and makes scant use of measure theoretical results or theorems to enable or facilitate the comparison of evaluation results. Thus evaluation research into AI - as critical as it is - resembles a sort of wild-west of largely incommensurable practices and incomparable results. And despite the avalanche of research into AI risk, we have no general theory of measurement for AI that sets standards and benchmarks for measurement practice and seeks to unify the disparate threads of evaluation and alignment into a more cohesive whole. This extended abstract seeks to address this by proposing an outline of what a measurement theory of artificial intelligence (MTAI) would look like - and why it would be worthwhile. 

\newpage
In this extended abstract we outline the motivation and characteristics a formal measurement theory for artificial intelligence.  Our central assumption is that measurement practice and methodology determines what we can say about AI systems, how we can compare them, and which quantitative risk tools we can soundly invoke.  Consequently, what AI \textit{is} - and how it may be controlled or ultimately aligned - is empirically contingent on how we choose to measure it. We argue that for these reasons, a comprehensive measurement theory for AI (MTAI) anchored in a synthesis of representation theories of measurement, metrology and underpinned by rigorous measure theoretic results is well motivated:

\begin{enumerate}
    \item \textit{Comparability and Cumulative Science}.  By defining constructs and scales with clear properties (e.g., scale type, invariance), an MTAI would facilitate commensurable comparisons across models, tasks, and research groups. This would help progress the field beyond leaderboard driven evaluation to a cumulative understanding of AI phenomena grounded in ``AI observables'' at different levels of the AI stack. A principled MTAI framework would (and ought to) enable commensurable evaluations across models, tasks, and research groups.  Without such a systematic framework (beyond benchmarks), comparative claims of superior reasoning, safety guarantees (even purported formal ones) and capabilities remain at risk of being limited at best.
    \item \textit{Standardisation}.  Disciplines such as reliability engineering, operations research, and quantitative risk analysis possess powerful mathematical tools—but these require variables that satisfy known scale types and invariance conditions.  An explicit MTAI would draw upon existing practices in metrology and risk-based sciences, providing for standardisation at the heart of other modern sciences. 
    \item \textit{Technical Engineering Benefits} From a technical perspective, an MTAI would enable (and arguably is a requirement of sector-wide) reliability engineering and quantitative risk analysis. By clearly specifying, for example, the characteristics of AI observables, their properties and interrelations (e.g. when they possess well-defined statistical and scale properties), control architectures can be more readily - and consistently - engineered into such systems. This is essential for evaluating and mitigating risks from frontier AI. It is also central for understanding - and engineering - interoperability of control processes across the AI stack to enable integration of, for example, potential mechanistically interpretable control at lower layers with prompt, training or infrastructure-based control. In this sense, measurement is the flip side of control - and so a well-developed MTAI would move AI control from an ad hoc set of initiatives towards a more formalised practice characteristic of other fields.
\end{enumerate}

The remainder of this abstract is set out as follows. First we discuss the problems and challenges faced in attempts to measure AI systems. Next, we expand upon rationales as to why an MTAI is needed. We then set out key characteristics of what an MTAI might encompass. Finally, we sketch out an example of how an MTAI would apply to different layers in the AI stack, from the physical to the sociotechnical.

\subsection{The Problem of AI Measurement}
Artificial intelligence systems can be stunningly complex. Contemporary large language models (LLMs) have hundreds of billions of parameters; reinforcement learning agents interact with dynamic environments via seemingly open-ended feedback loops; multi-agent systems evolve emergent behaviours not explicitly programmed by human engineers. Against this backdrop, it is important to ask what does it mean to measure AI? Consider the following criteria:
\begin{enumerate}
    \item \textit{Multiplicity of Attributes}. Unlike physical quantities—such as length or temperature—AI encompasses a variety of attributes: intelligence, capability, interpretability, fairness, robustness, alignment, etc. These are, at best, partially overlapping or ill-defined. Traditional measurement of physical magnitudes often relies on well-established axioms (e.g., length additivity). In AI, the theoretical basis for combining or even ordering these diverse constructs can be tenuous.

\item \textit{Evolving Systems}. AI is not static. It learns, adapts, and changes over time as new data, training, or fine-tuning processes are applied. This evolution complicates the usual notion of reliability in measurement. In classical terms, reliability requires that if a system does not intrinsically change, repeated measurements yield consistent values. But AI systems can shift internally in ways not directly observable.

\item \textit{Indirect Observations}. Much of what we want to measure in AI—capability, interpretability, potential risk—cannot be measured by simply reading off an instrument. Instead, we rely on indirect indicators: performance on benchmarks, resource usage logs, user satisfaction surveys, or inferences from system outputs. This is akin to psychometrics, where intangible constructs (e.g., intelligence, extraversion) must be inferred from test items or behavioural indicators.

\item \textit{Context Dependence}. AI performance or behaviour can vary dramatically with context (i.e., distribution shift, new tasks, or new user inputs). A system’s measured competence in one domain can fail to generalise. Measurement theory must accommodate conditional or contextual measurement—something rarely required in simpler physical measurement scenarios.
\end{enumerate}

Because of these challenges, contemporary research typically focuses on narrow, local metrics tailored to specific tasks. In practice, these local metrics work well within certain domains but provide no overarching unification. A measurement theory of AI would aim to provide formal methodological foundations across the AI stack.

\subsection{A History of Measurement}
What does it mean to measure artificial intelligence (AI)?  Historically, measurement theory has chiefly dealt with physical attributes (length, mass, time) or psychometric constructs (cognitive abilities, preferences).  AI, by contrast, introduces phenomena that range from low‑level hardware states to high‑level emergent behaviours (e.g., reasoning, creativity, or risk of misuse) which are difficult to measure and observe.  While evaluations of AI capabilities and risks are proliferating, they often rely on ad-hoc metrics and benchmarks that are not grounded in a coherent theory. This fragmentation problematises comparisons between models and architectures. It also complicates risk analysis and obscures the ways in which the choice of measurement protocol affects how AI is represented. \\
\\
The history of measurement can provide an initial guide as to how we ought to think about solving the measurement problem for AI. Attempts to formalise definitions of measurement have abounded since antiquity \cite{stanfordmeasurementscience}, often reflected in the classical distinction between sciences of quantity (number) and sciences of magnitude (geometry) of Euclid and Aristotle \cite{Aristotle1984,Euclid1908}. Concepts familiar to modern measurement theory and practice, including those of ratios (comparisons of magnitudes), have an ancient provenance. For Aristotle, quantities characterised by numbers could exhibit certain relations, such as equality or inequality, whereas qualities could be expressed in terms of degrees. Medieval scholarship saw further development of arithmetical concepts of magnitude (addition or subtraction \cite{Jung2011}) and the use of geometric and algebraic techniques to represent changes in magnitude and direction \cite{Clagett1968,Sylla1971}, including the quantitative characterisation of what were hitherto regarded as qualitative features \cite{Grant1996}. These early explorations of measurement formed the basis for the development of quantitative methods during the enlightenment, such as the principle of continuity (that change is a measure of degree, for both objects with extensive magnitude and intensional objects such as representational states) studied by Leibniz, Kant and others \cite{Kant1787,Jorgensen2009,Diehl2012}. Importantly they informed the development of early calculus and mathematical formalism which enabled the measurement of how objects and systems change over time. These developments culminated in the birth of the modern scientific practices of measurement in the 19th century \cite{Darrigol2003,Michell1993} and, in particular, scholarship on the mathematical foundation of measurement. As we discuss below, the formalisation of measurement theory and practice had significant consequences. One of these was enabling relationships - mappings - to be drawn between ontologies and objects in explanatory ways. Thus colour and hue, often a go-to example of qualia, became explainable in quantitative terms of spectra; foreshadowing later 20th century formalisations of measurement in measure-theoretic terms. It is also of relevance to the measurement of AI. For example, attempts to infer psychological qualities of properties of AI, such as dishonest, collusive, power-seeking or scheming behaviour \cite{Carlsmith2023scheming,Greenblatt2023aicontrol,Korbak2025sketch}.
\\
\\
Concurrently, advances in measurement technology saw the emergence of modern metrology with its focus upon units of measure and standardisation, such as the meter, kilogram, second, etc. This standardisation was guided by axiomatic principles (e.g., length is additive, masses combine linearly) and validated through empirical practice (e.g., repeated calibrations, consensus standards). In the 19th and 20th centuries, psychological sciences sought to define and measure intangible latent observables such as intelligence, personality, or attitude. They developed sophisticated measurement theories (e.g., Item Response Theory, factor analysis) to define scales that represent unobserved latent traits, many of which are being applied in how we think about the behaviour of AI systems. Across the AI evaluation landscape, we see parallels with all three traditions: from classical metrology’s emphasis on standardisation, from psychometrics’ approach to intangible constructs, and from modern statistics’ emphasis on data-driven model fitting. Yet AI measurement remains nascent and without the type of integration of formal methodological underpinnings that characterise physical and psychological sciences.  

% \subsection{Mathematical Formalism of Measurement}

\subsection{Does AI need Formal Measurement Theory?}
One immediate - and important - response to any MTAI proposal is to ask: why do we need a formal theory of measurement? Can we not simply rely upon typical ways we observe, assess and statistically measure any phenomena? After all, is this not what would be done in practice anyway? Answering these questions - akin to what is \textit{sui generis} about any MTAI is important. A few motivating reasons are set out below:
\begin{enumerate}
    \item \textit{Science of measurement}. Firstly, AI risk ought to be grounded in reproducible, robust and reliable scientific practice. Measurement is the cornerstone of empirical science. Without well-founded measures, it is difficult to accumulate replicable knowledge. For instance, if one research group claims an AI system is more interpretable or apparently robust than another, but uses an unvalidated ad-hoc metric, we cannot systematically compare results across studies or time. A formal measurement theory fosters consistency, reproducibility, and cumulative progress.
    \item \textit{Benchmarking and Comparison}. AI is already rapidly being integrated within  organisations, institutions and civil society. To assess the benefits from an economic perspective - and adjudge its risks - stakeholders reqiure a means of fair and meaningful comparison. This requires a rigorous, reproducible and testable means of answering questions like which system has stronger out-of-distribution generalisation? Or, which is safer or less prone to adversarial attacks? A shared measurement framework would allow measurement of such attributes in a comparable manner and interpretation of results similarly.
    \item \textit{Regulatory and Safety Concerns}. As AI is integrated into high-stakes scenarios—medical diagnosis, autonomous vehicles, large-scale financial trading—regulators, legislators, and auditors require transparent, standardised ways to measure compliance, reliability, or risk levels. Absent robust measures, oversight is haphazard at best.
    \item \textit{Ethics and Governance}. Many ethical frameworks for AI revolve around reducing harm, bias, and unfairness. These concepts need operationalisation in numeric or at least ordinal terms to enforce ethical guidelines or hold systems accountable. A formal measurement theory could build upon existing technical literature (such as in fair machine learning) that provides a model for how statistically-driven approaches can inform interpretation, enabling observable features of systems to be systematically tested for invariance across different populations or use cases. 
    \item \textit{Extrapolation and Forecasting}. In AI safety, we often worry about potential future capabilities or emergent behaviours. Valid measurement frameworks may help us detect early signs of emergent phenomena, monitor them quantitatively, and update risk assessments accordingly. A coherent approach to measurement may better equip us to forecast or anticipate transitions in AI’s competence or influence.
    \item \textit{Alignment with Foundational Measurement}. Establishing rigorous representational criteria can clarify when a given numerical scale truly captures an empirical structure—especially for intangible AI constructs (e.g., understanding, reasoning, intent). Doing so reduces the risk of ambiguous or misleading claims about an AI’s level or score of some property.
\end{enumerate}
There are also normative reasons why a specific measurement theory for AI ought to be developed. Firstly, there is benefit to researchers and developers of AI who would gain more consistent internal metrics for debugging and improving AI models. Moreover, a standardised measurement theory would facilitate cross-laboratory replication and help unify scattered benchmarking frameworks. Secondly, measurement theory would feed into standardisation (as it does in other engineering disciplines). Clear measurement standards would make it easier to compare AI products, ensure contractual obligations, and comply with emerging regulations. A robust and standardised set of measurement protocols may also assist in AI regulation and regulation of AI infrastructure, especially for AI agents \cite{chan2025infrastructure}. A rigorous foundation in measurement theory may also enable a more robust way to define and situate levels of AI oversight or licensing thresholds based on validated measures (akin to safety standards in other industries). In general, doing so will enable more evidence-based decisions regarding AI - helping delimit a more rigorous appreciation of what AI systems can and cannot do - an improvement by comparison with reliance upon heuristic judgments (which is especially important when the stakes are high). 

% Measurement theory also has an important role in delimiting what AI can and cannot do. It would form the cornerstone of rigorously comparable theories of AI capabilities and risk. Moreover, the standardisation of measurement would also enhance initiatives aimed at establishing trust and reliability of AI systems. [A well-articulated theory of AI measurement invites philosophical inquiry into machine agency, cognition, and intelligence, while also enabling robust experimentation in labs. It could spur new methodological innovations bridging the gap between classical measurement theory, large-scale data analysis, and advanced computational modelling.]

%%%%%%%%%%%

\section{What would a measurement theory for AI entail?}
\subsection{Normative Objectives}
If we accept that the need for an MTAI is well motivated, the question is then what form or forms it may take. Measurement as a practice is diverse and expansive in other sciences, but it is also standardised across the sciences through metrology and provided a rigorous technical underpinning by, for example, the mathematics of measure theory. What is measured across the sciences also differs, combining a synthesis of fundamental (direct) - such as metrological magnitudes - and derived (indirect) measurement - such as latent psychological or psychometric properties, or economic phenomena. Thus the question the shape and form of MTAI is potentially wide and varying. Moreover, it is important that we ask whether there is anything \textit{sui generis} about a MTAI? Is there anything new required of an MTAI beyond a mere application of existing methods? Our view, which we expand upon below, is that an MTAI requires a synthesis of different approaches to measurement across \textit{both} the physical and social sciences, depending on where in the AI stack measurement is performed (or control sought). Thus measurement of physical hardware signals (compute degree, circuit properties) may be metrologically-based in, while measurement of AI `behaviour', such as whether a model gives rise to purported scheming, collusive or nefarious behaviour may have more in common with psychometric evaluation techniques. In each case, it is reasonable to argue that a rigorous MTAI would satisfy the following:
\begin{enumerate}
    \item \textit{AI Observables and Foundations}. It would set forth foundational definitions of AI-related constructs (capabilities, reliability, interpretability, resource usage, etc.) in the form of \textit{AI observables}.
    \item \textit{Standardisation}. It would seek to standardise how we characterise or identify the AI stack. AI models vary considerably, as do their architectures, but an MTAI could in principle seek to identify a modular set of building blocks (such as those set out in the AI stack we utilise in this abstract) according to which AI systems in general may be characterised: they will all have a physical layer, for example; they will (almost always) comprise a software layer; they will always be situated within other `non-AI' architecture of infrastructure, and so on.
\item \textit{Formal Methodologies}. An MTAI would set out formal measurement practices of AI, including formal criteria for when numerical scales meaningfully represent these constructs. To this extent it ought to seek integration with well-established metrological and scientific practices of measurement, anchored in standardisation and statistical techniques.
\item \textit{Analysis of how AI evolves}. An MTAI needs to not only identify AI (characterise it), but also allow us to reason about how AI varies over time and response to perturbations (endogenous or exogenous). To this end, an MTAI ought to specify methods of invariance: transformations that preserve meaning, comparisons and AI observables across different AI systems.
\item \textit{Mathematical Rigour}. An MTAI must be standardised and ought to, to the extent possible, be grounded in underlying theories, such as measure theory, which allow us to reliably reason about AI observables, how they are formed, how they scale, vary, converge and so on. This does not mean all AI systems and all layers in an AI stack will be identical, nor that, for example, complex emergent capabilities of AI systems can be simply rendered using a single (or simple) measure. But it would provide a better statistical and empirical grounding to handle uncertain or evolving categories (e.g., new emergent capabilities). And in cases where we cannot identify (or it is impossible to identify) formal guarantees or mathematical underpinnings of AI observables, rigorously establishing this impossibility or difficulty itself has benefit, allowing us to catalogue in finer detail those properties of AI systems which are harder to measure - and therefore harder to control in principle. 
\end{enumerate}
Thus an MTAI would be analogous to classical measurement theory but adapted to the technical stack of AI, recognising that we do not always know the right units or categories a priori, but with a systematic method for how we would create or posit them (rather than arbitrarily deciding to anoint a new capability, risk class or effect). 

\subsection{Synthesising existing measurement theories}
In this abstract, we only sketch out a number of key principles and desiderata of an MTAI. But we argue that any MTAI ought to involve a synthesis of existing measurement practices. As noted above in our potted history of measurement - the art and practice of measurement is vast, extensive and varying. But we do have considerable standardisation. For example, modern empirical practice is underpinned by statistical techniques that themselves are given mathematically rigorous underpinnings via measure theory, analysis and other fields. This does not mean that most - or even many - measurement practices directly require measure theory, or that practitioners have a background in or require an understanding of the underlying mathematics of measure theory and analysis. But it does mean that, in effect, the statistical practices and standards adopted in experiments can be interrogated for rigour, even if the assumptions upon which experiments proceed may be debated. Modern measurement is a cumulative synthesis of different approaches to measurement that have developed in tandem with technology and theory over the years. MTAI would too reflect this synthesis. To this end, we argue that an MTAI could - and ought to - be based upon three pillars of representation theory, measure theory and metrology with psychometrics. We sketch out the relevance of each below (in other work we formalise these concepts and their interrelation in more detail):
\begin{enumerate}
    \item \textit{Representational Theory of Measurement (RTM).} RTM, formalised in considerable detail by Krantz, Suppes and others \cite{krantz_foundations_1971, suppes_foundations_1989} seeks to derive an axiomatic foundation for measurement. It defines measurement as the construction of a homomorphism from an \textit{empirical relational structure} (e.g., the set of AI systems with a relation `is at least as capable as') to a \textit{numerical relational structure} (e.g., the real numbers with `$\ge$'). The axioms of RM (e.g., transitivity, monotonicity) determine whether a meaningful scale can be constructed and what its properties (e.g., ordinal, interval, ratio) are. This gives us a formal language to define what constructs like capability or fairness must satisfy in order to be measurable.

    \item \textit{Measure Theory.} Measure theory \cite{tao2011introduction} would serve as the unifying mathematical language for MTAI. It allows us to define the space of AI states ($\Omega$), the set of observable events as a $\sigma$-algebra ($\mathcal{F}$), and probability distributions as measures ($\mu$) on that space. An `AI observable' could then constitute a measurable function $\phi: (\Omega, \mathcal{F}) \to (\mathbb{R}, \mathcal{B}(\mathbb{R}))$. Other results (such as Carath\'eodory's theorem) could be leveraged in order to provide formal guarantees in particular circumstances. This mathematical grounding would provide a more rigorous underpinning for probabilistic statements related to AI systems, statistical modeling, and definitions of expected performance or risk.

    \item \textit{Metrology and Psychometrics.} Metrology and psychometrics could be adapted for any MTAI:
    \begin{itemize}
        \item \textit{Metrology} is the science of modern measurement. It governs the direct measurement of physical quantities (e.g., power in watts, time in seconds) \cite{JCGM2012}. Metrological standards set out methods for calibration and traceability to established standards (e.g., SI units), which would ensure that AI-related measurements of the hardware layer, for example, are robust and comparable. Modern standards connected to metrology and statistics also set out practices for empirically valid experiments which could be adapted or drawn upon for MTAI.
        \item \textit{Psychometrics} provides a potential method for indirect measurement of intangible and latent constructs \cite{NunnallyBernstein1994,borsboom_measuring_2005}. Thus to measure interpretability or alignment, we could adopt like Item Response Theory (IRT) and Factor Analysis to build statistical models that infer a latent trait from a pattern of observable behaviours (e.g., performance on a suite of evaluation tasks).
    \end{itemize}
\end{enumerate}
\subsection{Direct and Indirect Measurement}
Any MTAI would also draw upon formal methodological distinctions regarding how explicit and latent observables of AI systems ought to be measured. Measurement is typically classified (as per Campbell) into \textit{fundamental} (direct) measurement and \textit{derived} (indirect) measurement \cite{campbell_physics_1920}. These distinctions are central to MTAI as, depending upon where in the AI stack measurement and control is sought, the nature of what is being measured will differ, being more or less direct depending on location:
\begin{enumerate}
    \item \textit{Fundamental (direct) measurement} may include (i) \textit{physical instrumentation}, such as voltage, energy consumption, bit-error rates. These can be read off instruments with known calibration standards; and (ii) \textit{model parameters} during various training stages, which may be directly observable (e.g. reading the numerical values of weights). Direct measurement typically yields raw data about AI’s internal states or resource usage. However, the meaning or significance of these raw measurements for AI capabilities or behaviours may not be directly clear without further data or interpretation. This represents one of the challenge of \textit{mechanistic interpretability} which represents an empirical attempt to measure AI systems in order to infer certain properties or behaviours arising from their underlying computational substrate or edifice.

\item \textit{Derived (indirect) measurement} may include (i) \textit{behavioural proxies}, such as success rates on tasks, user satisfaction scores, engagement metrics; (ii) \textit{correlated indicators}, measuring the text outputs of a model to infer changes in internal mood or reasoning style; or (iii) \textit{resource indicators}, such as energy consumption or carbon footprint as a proxy for computational load or training intensity.
\end{enumerate}
In addition, existing methods such as RTM can be drawn upon. An examples includes \textit{additive conjoint measurement}, which shows how fundamental, interval-scaled measures can be constructed for multiple attributes simultaneously by observing their joint effects \cite{luce_simultaneous_1964}. This may provide a formal means for creating robust scales for complex AI phenomena which involve the trade-off between performance and efficiency.

%%%%%%%%%%%%

\subsection{Characteristics for AI Measurement}
An important feature of any measurement theory is to understand and provide a procedure for enumerating exactly what we are measuring i.e. the phenomena of interest being measured. In a domain suuch as AI, we may not yet know the relevant categories. Theoretical approaches that form the basis of measurement practice can broadly be classified into formal (or axiomatic) and statistical (or empirical). We outline a few key issues regarding the application of axiomatic as opposed to empirical approaches to any MTAI below.

\subsubsection{Axiomatic (Formal) Measurement}
Classical axiomatic measurement theory begins by defining:
\begin{itemize}
    \item A set of empirical objects X.
    \item A set of relations $\{R\}$ on $XRX$, such as is at least as large as, or is the concatenation of.
\item A function $\phi:X \to \mathbb{R}$, $X \to \Real$ which represents those relations numerically.
\end{itemize}
If these relations satisfy certain axioms (e.g., completeness, transitivity, monotonicity), representation theorems guarantee that $\phi$ is well-defined and that certain uniqueness or invariance properties hold (e.g., only certain transformations may be permitted for an interval scale).
In AI, we might define an empirical relation for AI system A is at least as capable as AI system B on task set T. If we ascertain that this relation is transitive, consistent across repeated tests, and so forth, we might prove an interval or ordinal scale exists. However, many open questions remain: does the domain unify across tasks? Does partial ordering break down if A surpasses B in some tasks but not others? 
Formal or axiomatic measurement theories usually proceed by specifying a formal ontology \cite{Guarino_Oberle_Staab_2009} of what is (or exists or is to be observed) together with how those objects relate (a theory of relations):  \begin{enumerate}[label=(\roman*)]
    \item \textit{identification} - they identify plausible constructs or phenomena. For AI these may be system performance, reliability, interpretability, risk of undesirable behaviour, resource usage, etc; 
    \item \textit{relations} - relations among the objects are interrelated, often in the form of minimal relational axioms. An example might be, if we want a scale for AI capability, we might assume transitivity (if model A is more capable than B, and B is more capable than C, then A is more capable than C). 
\end{enumerate}
This approach is a direct application of the Representational Theory of Measurement (RTM). The goal of the axiomatic method is to explicitly define the \textit{empirical relational structure} for a given AI attribute. The axioms (transitivity, monotonicity, etc.) are postulates about this structure. A representation theorem then proves that if an empirical system satisfies these axioms, a homomorphism to a specific numerical relational structure (e.g., the real numbers with the relation $\ge$) exists and is unique up to a certain class of transformations \cite{krantz_foundations_1971}. The type of scale (ordinal, interval, ratio) is determined by the strength of these axioms.
\\
\\
In classical (axiomatic) measurement theory \cite{krantz1971foundations,luce2007foundations}, one starts by formalising empirical structures (ordering, concatenation, threshold relations) and seeks representation theorems. A \textit{representation} is a numerical function $f$ that preserves the empirical relations. We then seen \textit{uniqueness} and \textit{invariance} conditions, namely constraints on transformations that preserve meaning (e.g., strictly increasing transformations for ordinal scales, affine transformations for interval scales, etc.). In such case (in analogy with other mathematical fields, such as representation theory, we can denote the representation \textit{faithful} such that we can study the underlying phenomena by studying properties of its representation. Practically speaking an axiomatic theory may facilitate the following:
\begin{itemize}
    \item \textit{Ordinal comparison}. We use axiomatic theories to define an ordinal scales in order to compare AI system properties or attributes, such as capability. That is, comparisons of AI capabilities rely upon trusting pairwise comparisons on tasks (A outperforms B) which is itself conditional on each system having sufficient properties in common (or in more formal treatments, something resembling a univariate measure).
    \item \textit{Ratio scales}. Theory can be used to construct ratio scales if we are faced with zero or multiplicative comparisons e.g. between computational cost or energy of different AI systems. This is important particularly as we consider how the transformations of underyling AI systems may transform underyling capabilities, including whether, for example, certain capabilities scale or transform in the same way as some property or feature of an underlying model.
\end{itemize}
These formal scales would require that the underlying phenomena satisfy certain axioms (e.g., transitivity, monotonicity). Violations or complexities (e.g., a model that’s better at some tasks but worse at others) may require more complex or multi-dimensional approaches. A core tenet of measurement theory is that the type of measurement scale is defined by the transformations that leave the structure of the scale invariant \cite{stevens_theory_1946}. Understanding the scale type of an AI observable is crucial because it informs which mathematical operations and comparisons are useful. We can classify potential AI measurements according to classical measurement hierarchy, as shown in Table \ref{tab:scales}.

\begin{table}[h!]
\centering
\caption{Stevens' Scale Types Applied to AI Measurement}
\label{tab:scales}
\begin{tabularx}{\textwidth}{@{}l>{\raggedright\arraybackslash}X>{\raggedright\arraybackslash}X@{}}
\toprule
\textit{Scale Type} & \textit{Permissible Transformations} & \textit{AI Measurement Example} \\
\midrule
\textit{Nominal} & Any one-to-one mapping & Classifying models by architecture (e.g., `Transformer', `CNN', `RNN'). T\\
\addlinespace
\textit{Ordinal} & Monotonically increasing function ($f(x) \ge f(y)$ iff $x \ge y$) & Ranking models on a leaderboard (e.g., HELM, GLUE). We know model A is better than B, but not by how much. \\
\addlinespace
\textit{Interval} & Affine transformation ($ax+b$, with $a>0$) & A hypothetical `capability' score from an Item Response Theory model, where the zero point is arbitrary. We can meaningfully compare the intervals between scores. \\
\addlinespace
\textit{Ratio} & Multiplicative scaling ($ax$, with $a>0$) & Measuring training time, energy consumption (kWh), or parameter count. These have a true, non-arbitrary zero point, so ratios are meaningful. \\
\bottomrule
\end{tabularx}
\end{table}

\subsubsection{Modern Statistical / Psychometric Paradigms}
Realistically, we often do not start with a crisp set of axioms for intangible constructs like interpretability or general intelligence. Instead, we gather empirical data-system performance logs, user ratings, or bench test outcomes—and apply statistical models (factor analysis, latent trait modelling, or Bayesian hierarchical inference) to detect patterns. This second approach is empirical, involving a statistically-driven approach, albeit with a theoretical underpinning in modern statistical methodologies. Modern measurement theory (e.g., factor analysis, item response theory, cluster analysis) applied statistical methods to classify, essentially as a means of revealing the ontology via statistical analysis. Thus statistics may reveal latent dimensions or categories from large datasets of AI behaviour. Doing so may facilitate discovery of new constructs (e.g., emergent theory of mind capacity in large language models). Statistical methods are at the heart of empirical methods in the sciences and key for MTAI.  In each case, the complexity of modern AI systems means that purely axiomatic frameworks are too rigid and may often lack utility. Statistical methods may be utilised to instead frame measurement in different ways:
\begin{itemize}
    \item \textit{Distributions and Measure}. Formal statistical methods rely upon a mathematical underpinning of measure theory \cite{tao2011introduction}. Measure theory provides the theoretical tools which enable us to reason and calculate about AI systems using probability, sampling and statistical methods. This may include whether how the structures or characteristics inferred from measurement and observation scale, transform or converge as AI systems scale.
    \item \textit{Latent variables}. AI system attributes e.g. capability or bias may be considered a latent dimension that influences outputs, behaviour, task completion or scenarios across variations. Latent variables may be estimated using parametric modelling (e.g. logit, probit or neural networks).
    \item \textit{Empirical classification}. Ontological classification is derived from empirical methods applied to data (including measurement itself). This may include factor analyses, manifold learning, or clustering methods to detect AI behaviours or trends in AI system-evolution.
    \item \textit{Measurement invariance}. Often we are interested in scale-free or invariant measures of AI attributes. Identifying measurement invariants e.g. whether the same classes appear at different scales, or what they are contingent upon (e.g. attributes appearing in larger-dimensional but not smaller-dimensional systems), or where properties do or do not observably subsist across different contexts, training sets, or user populations is a means of classifying robustly measurement ontology. The absence of invariance in such classifications means that we may require additional classes or contingencies (e.g a scale-dependent measurement ontology), which is typical (we do not measure the same phenomena at the quantum scale versus the human scale, for example). 
\end{itemize}
This methodology aligns with what contemporary debates regarding model-based accounts of measurement \cite{frigerio_outline_2010, boumans_measurement_2015}, where measurement is not the direct reading of a pre-existing value, but the coherent assignment of a value to a parameter within a statistical model of the measurement process. This perspective introduces a crucial distinction between \textit{instrument indications} (e.g., a raw score on a benchmark) and \textit{measurement outcomes} (the inferred latent trait, like capability) \cite{jcgm_international_2012}. Statistical models serve as the inferential bridge between the two, and would provide, for example, the confidence to more rigorously reason about apparent latent capabilities of AI systems, such as psychologically-described behaviour of scheming, deception or claims regarding simulation of human cognition \cite{binz2024centaur}.

\subsection{The Role of Mathematics}
Both axiomatic, statistical and measure theoretic approahces to MTAI highlight the need for MTAI to be mathematically formal. This includes that MTAI ought to facilitate the identification and specification of the following aspects of any AI system to which it is applied:
\begin{itemize}
    \item \textit{Empirical Systems and Relations}. For example, let $\mathcal{S}$ be the set of AI systems under investigation and $\succeq$ a binary relation capturing `is at least as capable as'. Potential tasks or data sets can be included as parameters, so we might define a relation $\succeq_T$ for tasks $\mathcal{T}$. 
    \item \textit{Representation Functions}. A function $\phi_T: \mathcal{S} \to \mathbb{R}$ might represent how each system $s \in \mathcal{S}$ performs on task $T$. Ideally we would seek $\phi_T$ to (homomorphically) preserve $\succeq_T$ in a manner consistent with an ordinal or interval scale i.e. $s_1 \succeq s_2 \implies \phi_T(s_1) \leq \phi_T(s_2)$.
    \item \textit{Additivity or Concatenation}. If measuring resource usage, we might want an additive or at least monotonic combination property. For instance, the resource usage of combined sub-tasks or concurrent modules might be represented by numeric addition or a known function.
    \item \textit{Probability Distributions}. In a more advanced formulation we can incorporate probabilistic representations to capture noise, random seeds, or environment stochastic. This leads us to define random variables $X_{s,T}$ representing system $s$'s performance on task $T$.  The measurement problem is then to define a scale that captures the distributional relation $X_{s_1,T} \succeq X_{s_2,T}$ in some robust sense (e.g., first-order stochastic dominance or other partial ordering of distributions).
    \item \textit{Uniqueness and Invariance}. We may be interested in which set of transformations $\tau$ on $\mathbb{R}$  preserve the meaning of our scale? If we only assume an ordinal structure, then $\tau$ must be strictly increasing. If we assume an interval structure, then $\tau$ must be affine i.e. $\tau (x) = ax + b$. Establishing what type of scale is appropriate for each AI attribute is a major theoretical challenge.
\end{itemize}
These preliminary mathematical sketches illustrate how the building blocks from classical measurement theory could be adapted to AI. However, the scope of application is far broader, as AI systems are not single-variable but multi-layered and multi-dimensional, necessitating a mosaic of partial or multi-attribute measurement approaches.

\subsection{Challenges}
Despite the potential benefits and the overarching frameworks, numerous obstacles complicate the pursuit of any MTAI: 
\begin{enumerate}
    \item \textit{Faithfulness}. The faithfulness of representations is often problematic, with fundamental uncertainty existing about how and whether we can faithfully represents certain AI behaviours or characteristics in crisp circuit formalism for example.  
    \item \textit{Non-Transitive or Incomplete Orderings}. An AI system may excel at some tasks while lagging on others; we might not have a universal better or worse relationship across all tasks. Thus, a single linear scale for capability might not exist. Partial or multi-dimensional orderings might be necessary.
    \item \textit{Context-Dependent Shifts}. AI systems frequently face distribution shifts: a model trained on one domain fails in another. A measurement that is stable in domain A might lose validity in domain B, complicating any universal scaling approach.
    \item \textit{High-Dimensional and Continuous Evolution}. Because AI systems can continuously learn or fine-tune, the property being measured can shift in real time. Traditional measurement presupposes that the object’s measured property is at least relatively stable while measuring it.
    \item \textit{Black-Box or Proprietary Systems}. If many modern AI systems are proprietary black boxes, obtaining direct measurements from within (e.g., internal weight states or resource usage logs) might be impossible. We might only have access to outputs, which severely limits measurement approaches or forces reliance on behavioural tests reminiscent of psychometric item sets.
    \item \textit{Socio-Technical Interactions}. AI does not exist in isolation: it interacts with users, organisations, societies. This environment shapes how we interpret reliability, safety, or trustworthiness. A purely system-centric measurement might miss crucial aspects of emergent socio-technical phenomena.
\end{enumerate}
These obstacles underscore the need for a unified perspective that can situate each difficulty at the layer of abstraction where it arises while still permitting rigorous comparisons across layers. By reframing AI systems as a stratified stack, with each tier endowed with its own observables, instrumentation requirements, and validity conditions, we can tame the apparent heterogeneity of metrics and methods. The following section formalises this idea as a measurement stack, detailing how layer-specific measures compose into a coherent, end-to-end framework for evaluating and controlling frontier AI. We sketch an example of such a stack below.
%%%%%%%%%%
\subsection{Measurement Stack}
Given the enormous variety of phenomena designated as or as relating to AI, a measurement theory must be sufficiently versatile for use across multiple layers of the AI stack \cite{perrier2025out}, not simply at those layers for which we have well-established measurement protocols (such as the hardware layer). To address this lack of unity and coordination in how AI is measured and synthesise the three main pillars of measurement theory identified above,  we propose a layered `measurement stack' that decomposes an AI system into distinct levels of abstraction, each suited to different measurement paradigms. These layers present different measurement targets, methods, and complexities. We sketch out a proposed (by no means complete or mutually exclusive) way of conceptualising the AI stack and its measurement below:

\begin{enumerate}
\item \textit{Physical Layer}. Observables may include: circuits, transistors, voltage levels, chip architecture, and power supply.
Here, classical engineering measurements and metrology (with traceability to SI units) apply relatively straightforwardly. Energy usage and thermal dynamics can be measured with established instrumentation.
 MTAI would ideally incorporate recognised engineering measurement standards to ensure alignment with the rest of science and technology.
 \item \textit{Systems Layer}. Observables may include: operating systems, compilers, libraries, resource scheduling, concurrency.
Measurable phenomena at this layer may also include latency, throughput, reliability of system calls, concurrency overhead, etc.
 These metrics can feed into higher-level indicators of resource usage efficiency or real-time operational constraints.
\item \textit{Algorithm/Model Layer}. Observables may include: neural network weights, knowledge graph edges, decision rules in symbolic AI, or policy networks in reinforcement learning. Many types of measurements at this level are more abstract, including the distribution of weights, gradient magnitudes, or topological structure of layers. 
 MTAI would seek to define how these quantities relate to higher-level conceptual constructs e.g., overfitting, learned representations, internal states predictive of future performance.
\item \textit{Task/behaviour Layer}. The observables at this layer include the direct outputs or actions in response to inputs of an AI model such as: classification labels, textual responses, recommended decisions. Metrics of relevance for MTAI may include performance or accuracy metrics, reliability across distribution shifts, or user satisfaction.  A more classical stance might see these as directly measurable outcomes, but deeper constructs (capability, trustworthiness) remain latent and require interpretive frameworks or test items (benchmarks, adversarial examples).
\item \textit{Contextual/Emergent Layer}. Emergent phenomena like cooperation in multi-agent systems, theory of mind, or the system’s alignment with human values may also form a class of observables. Here, intangible constructs must be inferred from possibly numerous indicators. Formal definitions remain in flux.  This layer most strongly parallels psychometrics: intangible or ephemeral properties gleaned from patterns of behaviour. A thorough measurement theory must clarify the axioms or assumptions needed to treat these emergent traits as measureable in the first place.
\end{enumerate}
Because each layer has distinct measurement challenges, the scope of a general AI measurement theory must be modular. It should allow for different sub-theories or measurement protocols at each layer, yet offer overarching principles for how these sub-measures might combine or be related to higher or lower layers. An MTAI would seek to rigorously standardise the AI stack in a modular way, thereby enabling comparison of different AI configurations.

\section{Examples}
In our final section, we set out a few tentative examples of how the synthesis of existing measurement techniques might set an outline for an MTAI. 

\subsection{Meta-Axioms}
One approach towards formally underpinning measurement of AI from an axiomatic perspective might be the application of classical RTM via the construction of meta-axioms of measurement. Here, we would require that every measurement proposition \(P\) would be declared together with an
ordered quadruple:
\[
  \bigl(
    \text{Empirical System } \EmpSys,\;
    \text{Numerical System } \NumSys,\;
    f : \EmpSys \!\longrightarrow\! \NumSys,\;
    \UniqGrp
  \bigr),
\]
where
\begin{enumerate}[label=(\roman*)]
  \item \(\EmpSys\) is the set (or structured space) of empirical objects or
        events to which \(P\) refers;
  \item \(\NumSys\) is the codomain endowed with its canonical order
        \((\NumSys,\ge)\) or richer algebraic structure;
  \item \(f\) is a \emph{representation homomorphism} preserving all empirical
        relations of interest (\(\forall a,b\in\EmpSys\):
        \(a \preceq_{\EmpSys} b \iff f(a)\ge f(b)\));
  \item \(\UniqGrp \le \mathrm{Aut}(\NumSys)\) is the \emph{uniqueness group}
        consisting of every scale transformation
        \(\tau : \NumSys \!\to\! \NumSys\) that leaves the empirical
        information content of \(f\) invariant
        (\(f\) and \(\tau\!\circ f\) represent the \emph{same} measurement).
\end{enumerate}
Such a meta-axiom would be designed to force explicit declaration of underlying assumptions, requiring enunciation of the empirical domain being measured, the numerical structure being mapped to, what transformations preserve the measurement meaning and the scale. The benefit of an approach such as this is that it can more clearly identify when, for example, an AI property is being evaluated or benchmarked, the extent to which that property is amenable to mathematical formalisation (and therefore rigorous comparison). 

%%%%%%%observables
\subsection{AI Observables}
Another potential approach (which we expand upon in our more detailed work) is in offereing a definition of an \textit{AI observable} and justifying why measure theory is useful for considering how we frame and reason about AI observation. to this end, one potential approach of MTAI is to define AI observables. An \textit{AI observable} would be a measurable function $\phi: (\Omega, \mathcal{F}) \to (\mathbb{R}, \mathcal{B}(\mathbb{R}))$ where:
\begin{itemize}
\item $\Omega$ represents the space of all possible AI system states or configurations
\item $\mathcal{F}$ is a $\sigma$-algebra on $\Omega$ representing the collection of events we can distinguish through measurement
\item $\mathcal{B}(\mathbb{R})$ is the Borel $\sigma$-algebra on $\mathbb{R}$
\item $\phi^{-1}(B) \in \mathcal{F}$ for all $B \in \mathcal{B}(\mathbb{R})$ (measurability condition)
\end{itemize}
Here $\Omega$ the fundamental ontology of an AI measurement theory—it specifies what exists that can be measured. This is not merely a technical detail but an important assumption about the nature of AI systems and their measurable properties. This formal definition allows us to frame a central philosophical question in AI measurement. Are these observables referring to real, mind-independent properties of the AI system, or are they convenient fictions defined by our measurement procedures? This is the classic realism versus anti-realism debate. A realist, for instance, would argue that a latent attribute like `capability' corresponds to an objective, structured property of the model \cite{swoyer_metaphysics_1987, borsboom_measuring_2005}. An operationalist might argue, for example, that the concept of capability \textit{is} synonymous with the set of operations used to measure it, such as performance on a specific benchmark \cite{bridgman_logic_1927}. 
\\
\\
Adopting this or a similar type of definition for AI observables may assist in helping reveal why certain proposed measurements of AI systems may be ill-defined: if the purported observable cannot be expressed as a measurable function on the system's state space, it lacks mathematical foundation. For instance, vague notions like ``understanding'' or ``consciousness'' must first be operationalised as measurable functions before they can serve as comparable observables. The benefit of this approach is that we do not require a final or universal attempt to define what are vague and often contested concepts - such as whether AI is conscious. Rather, what counts in an MTAI is whether we have a \textit{measurement procedure} for repeatedly and rigorously observing and measuring the purported phenomena. In this way, a rigorous MTAI may assist in moving forward debates by concentrating attention on the empirical practice of measurement rather than nebulous concepts.

\section{Conclusion}
In this extended abstract, we have sought to motivate and outline the form of a prospective measurement theory of artificial intelligence. As one of the most - if not the most - significant technologies of our era, ensuring a scientific approach to measurement beyond mere ad hoc evaluation or benchmarking is crucial to the understanding of AI systems, their evolution and importantly their control. An MTAI would contribute to achieving this objective. Our framework for MTAI adopts a position of \textit{methodological realism}. We hypothesise that stable, latent attributes of AI systems exist, and we use the tools of measurement theory---both axiomatic and statistical---to construct and validate representations of observables. The success of a general measurement model in providing consistent, coherent, and predictive results serves as evidence for the reality of the underlying construct. Our more extensive treatment of MTAI is to be set out in an upcoming work.

%%%%%%%%%%Synthesis

%%%%%%%%BIBLIOGRAPHY

\bibliographystyle{unsrt}
\bibliography{refs-new,refs-measurement,refs-measure,refs-aicontrol,refs-aisafety,refs-meas-new}

% \appendix

% %%%%%%%%%%%%%SUMMARY
% \section{Information theory}
% abc

% \section{Quantum measurement}

\end{document}